# Are Traditional Deep Learning Model Approaches as Effective as a Retinal-Specific Foundation Model for Ocular and Systemic Disease Detection?

Samantha Min Er Yew[1,2*], Xiaofeng Lei[3*], Jocelyn Hui Lin Goh[1,2,4*], Yibing Chen[4], Sahana Srinivasan[1,2,4], Miao-li Chee[4], Krithi Pushpanathan[1,2], Ke Zou[1,2], Qingshan Hou[1,2], Zhi Da Soh[4], Cancan Xue[4], Marco Chak Yan Yu[4], Charumathi Sabanayagam[4,5], E Shyong Tai[1,6,7,8], Xueling Sim[6], Yaxing Wang[9,10], Jost B. Jonas[4,9,11,12], Vinay Nangia[13], Gabriel Dawei Yang[14], Emma Anran Ran[14], Carol Yim-Lui Cheung[14], Yangqin Feng[3], Jun Zhou[3], Rick Siow Mong Goh[3], Yukun Zhou[15,16,17], Pearse A. Keane[16,18], Yong Liu[3†], Ching-Yu Cheng[1,2,4,5†], Yih-Chung Tham[1,2,4,5†]

*contributed equally as first authors

†contributed equally as last authors

[1] Department of Ophthalmology, Yong Loo Lin School of Medicine, National University of Singapore, Singapore.

[2] Centre for Innovation and Precision Eye Health, Yong Loo Lin School of Medicine, National University of Singapore, Singapore.

[3] Institute of High-Performance Computing (IHPC), Agency for Science, Technology and Research (A*STAR), Singapore, Singapore.

[4] Singapore Eye Research Institute, Singapore National Eye Centre, Singapore, Singapore.

[5] Ophthalmology and Visual Science Academic Clinical Program, Duke-NUS Medical School, Singapore, Singapore.

[6] Saw Swee Hock School of Public Health, National University of Singapore, Singapore, Singapore.

[7] Duke-NUS Medical School, Singapore, Singapore.

[8] Precision Health Research, Singapore, Singapore.

[9] Beijing Visual Science and Translational Eye Research Institute (BERI), Beijing Tsinghua Changgung Hospital, Tsinghua Medicine, Tsinghua University, Beijing, China.

[10] Tsinghua Medicine, Tsinghua University, Beijing, China.

[11] Rothschild Foundation Hospital, Institut Français de Myopie, Paris, France.

[12] New York Eye and Ear Infirmary of Mount Sinai, Icahn School of Medicine at Mount Sinai, New York, United States.

[13] Suraj Eye Institute, Nagpur, India.

[14] Department of Ophthalmology and Visual Sciences, The Chinese University of Hong Kong, Hong Kong Special Administrative Region, China.

[15] Centre for Medical Image Computing, University College London, London, United Kingdom.

[16] NIHR Biomedical Research Centre, Moorfields Eye Hospital NHS Foundation Trust, London, United Kingdom.

[17] Department of Medical Physics and Biomedical Engineering, University College London, London, United Kingdom.

[18] Institute of Ophthalmology, University College London, London, United Kingdom.

**Conflict of Interest:** None of the authors have any conflicts of interest to disclose.

**Corresponding author:**

Dr. Yih-Chung Tham

**Abstract**

Background: RETFound, a self-supervised, retina-specific foundation model (FM), showed potential in downstream applications. However, its comparative performance with traditional deep learning (DL) models remains incompletely understood. This study aimed to evaluate RETFound against three ImageNet-pretrained supervised DL models (ResNet50, ViT-base, SwinV2) in detecting ocular and systemic diseases.

Methods: We fine-tuned/trained RETFound and three DL models on full datasets, 50%, 20%, and fixed sample sizes (400, 200, 100 images, with half comprising disease cases; for each DR severity class, 100 and 50 cases were used. Fine-tuned models were tested internally using the SEED (53,090 images) and APTOS-2019 (3,672 images) datasets and externally validated on population-based (BES, CIEMS, SP2, UKBB) and open-source datasets (ODIR-5k, PAPILA, GAMMA, IDRiD, MESSIDOR-2). Model performance was compared using area under the receiver operating characteristic curve (AUC) and Z-tests with Bonferroni correction (P<0.05/3).

Findings: For ocular diseases, fine-tuning on full datasets showed comparable internal performance between traditional models and RETFound (AUCs: 0.914–0.965 vs. 0.938–0.966). With smaller datasets, trends remained, except for DR (≤100 images per class) and glaucoma (≤400 images), where ResNet50 was inferior to RETFound (all P≤0.005), though SwinV2 remained comparable. External validations were consistent. Conversely, for systemic diseases, when fine-tuned with smaller datasets, RETFound consistently outperformed traditional models. Specifically, with 100 images, RETFound exceeded traditional models for detection of diabetes (RETFound's AUC: 0.648 vs. ResNet50: 0.612, SwinV2: 0.580; all P≤0.001), hypertension (RETFound: 0.705 vs. ResNet50: 0.634, SwinV2: 0.648; all P<0.001), and CKD (RETFound: 0.822 vs. SwinV2: 0.778, ViT-base: 0.743, all P≤0.005).

Interpretation: Traditional DL models are mostly comparable to RETFound for ocular disease detection with large datasets. However, RETFound is superior in systemic disease detection with smaller datasets. These findings offer valuable insights into the respective merits and limitation of traditional models and FMs.

**Research in context**

Evidence before this study: We searched PubMed for articles published in English from database inception until Jan 8, 2025, using keywords "RETFound", "traditional model", and "comparison". We found limited evidence on the comparative performance of RETFound and traditional deep learning model approaches in detecting ocular and systemic diseases using coloured retinal photographs. Prior studies focused primarily on fine-tuning RETFound's performance for single ocular disease outcomes and often lacked external validation. To the best of our knowledge, no previous study has conducted a comprehensive comparative evaluation of RETFound and traditional deep learning models across major ocular and systemic disease outcomes, utilizing multiple datasets and systematically accounting for varying fine-tuning scenarios, including small sample sizes in both proportional and absolute terms, within a single study.

Added value of this study: In this study, we conducted extensive fine-tuning experiments on RETFound, using varying sizes of fine-tuning datasets to simulate a broad range of data resource scenarios. These experiments were repeated on three other traditional deep learning models (i.e., supervised learning approaches utilizing ResNet50 and vision transformer backbones). To ensure the robustness of our findings, we performed a comprehensive head-to-head evaluation of the models' performance across multiple internal and external test sets, derived from diverse population-based studies and open-source datasets for various ocular and systemic disease outcomes. Our findings provide valuable insights into the limited evidence regarding the performance advantages of RETFound over traditional deep learning approaches, which may offer critical guidance on ongoing research for foundation models.

Implications of all the available evidence: Our study highlights the performance advantage of fine-tuning RETFound for oculomics tasks in low data resource settings compared to traditional deep learning approaches. However, as deep learning techniques continue to evolve and new foundation models emerge, further benchmarking evaluations using similar methodologies will be essential.

**Introduction**

Artificial intelligence foundation models (FMs) have emerged rapidly with notable use cases across several medical domains including cardiology[1], oncology[2], histopathology[3] and ophthalmology[4,5]. The core concept of FMs involves training deep learning (DL) models to extract and learn features and patterns from vast amounts of unlabeled medical data through pretext tasks, often pre-trained using self-supervised learning (SSL) approach[6]. These FMs are then fine-tuned and adapted for domain-specific downstream tasks, without requiring minimal labelled medical data. In contrast, commonly used traditional DL models typically adopt the approach of pretraining on natural image datasets (i.e. ImageNet), utilizing transfer learning followed by supervised learning (SL) to perform single, task-specific medical applications.[7]

Due to their large pre-training datasets and SSL approach, FMs demonstrate promising adaptability.[8,9] Released in 2023, RetFound is the first FM in Ophthalmology. RETFound was sequentially trained on 1.3 million natural images followed by ~900,000 colour fundus photographs (CFPs) and ~700,000 optical coherence tomography (OCT) scans, using a generative self-supervised learning (SSL) technique called masked autoencoders (MAE).[4] RETFound was subsequently fine-tuned and demonstrated to be effective when adapted for several downstream tasks including retinal disease diagnosis and systemic disease prediction.[4]

While the original RETFound study showcased promising potential of an FM-based pretraining approach, several knowledge gaps still remain. First, the comparative performance and utility of fine-tuning on RETFound relative to current commonly used traditional DL model approaches (e.g., SL on ResNet50, Vision Transformer), remains incompletely understood. Second, comparisons between these models and approaches in critical aspects such as label efficiency and computational resource requirements, and generalizability, have not been thoroughly explored. Hence, a more holistic evaluation is needed, encompassing a broad range of downstream tasks with extensive external testing.

To address these gaps, we aimed to evaluate and compare the performance of RETFound alongside traditional supervised learning DL model approaches, namely ResNet50, ViT-base, and SwinV2, on clinical tasks involving ocular and systemic diseases. These models were fine-tuned/ trained across varying dataset sizes and tested on multiple external datasets. This comprehensive evaluation may provide critical insights into the practical considerations of adopting RETFound fine-tuning over traditional DL model approaches, accounting for accuracy, label efficiency, and computational resource demand, and overall clinical applicability.

**Methods**

**Study Design**

**Figure 1** provides an overview of our study. We retrospectively evaluated and compared RETFound with three traditional DL models which included ResNet50[10], SwinV2[11] and ViT-base[12], across four ocular disease detection downstream tasks and three systemic disease detection tasks. These three traditional DL models were selected due to their current widespread use and representational capacity.[12-14] These features placed them as reasonable benchmarks in computer vision research and for performance evaluations across diverse tasks and datasets.[15,16] The ocular diseases detection tasks included disease-related visual impairment (VI), visually significant cataract, diabetic retinopathy (DR), and glaucoma, while the systemic disease detection tasks included the detection of diabetes, hypertension, and chronic kidney disease (CKD). Detailed disease definitions for each task are provided in **Supplementary Table 1.** These evaluations were conducted using varying sample sizes for fine-tuning RETFound or training the traditional models.

To accurately reflect the distinct technical optimization approaches, we clarified that fine-tuning specifically refers to the adaptation of pre-trained RETFound weights for specific tasks, whereas training applies to traditional models initialized with ImageNet-pretrained weights and subsequently optimized for each task. Nonetheless, for ease of reading, we used the term "fine-tuning" for both RETFound and traditional models in the subsequent results and discussion section.

**Datasets for ocular disease detection**

The number of images used for fine-tuning/training, as well as internal and external testing in each task, is summarized in **Supplementary Table 2**.

For detection of disease-related VI, the models were fine-tuned (for RETFound)/trained (for traditional models), and internally tested on the CFPs and clinical data from the Singapore Epidemiology of Eye Disease (SEED) study, which comprises of the Singapore Malay Eye study (SIMES), the Singapore Chinese Eye study (SCES) and the Singapore Indian Eye study (SINDI).[17] External testing for this task was conducted on datasets from the Central India Eye and Medical study (CIEMS)[18], the Beijing eye study (BES)[19], and the Chinese University of Hong Kong's Sight Threatening Diabetic Retinopathy study (CUHK-STDR)[20].

For detection of visually significant cataract, the models were fine-tuned/trained and internally tested on the SIMES dataset, with external testing performed on the SCES, SINDI and BES datasets.

For the detection of diabetic retinopathy, the models were fine-tuned/ trained and internally tested on the APTOS-2019 dataset (India)[21], with external testing on two other publicly available datasets: IDRID (India)[22] and MESSIDOR-2 (France)[23].

For the detection of glaucoma, the models were fine-tuned/trained and internally tested on the SEED dataset, with external testing performed on publicly available datasets from ODIR-5K (China)[24], PAPILA (Spain)[25] and GAMMA (China)[26].

**Datasets for systemic disease detection**

For the detection of diabetes, hypertension, and CKD, we fine-tuned/ trained and internally tested all models using the SEED study datasets with external testing performed on datasets from the BES, the Singapore Prospective study (SP2)[27,28] and the United Kingdom Biobank (UKBB)[29] datasets.

**Model pre-training**

The ResNet50 model was initialized with pre-trained weights derived through SL from the ImageNet-1k dataset, which contains 1.4 million labeled natural images across 1,000 categories. ViT-base and SwinV2 were initialized with pre-trained weights derived through SL from the ImageNet-21k dataset, consisting of 14 million labeled natural images across 21,000 categories. Transfer learning was subsequently applied in these three models.

In contrast, RETFound was initialized with SSL weights pretrained sequentially using MAE on ImageNet-1k, followed by 904,170 unlabeled CFPs and 736,442 OCT scans from the Moorfields Diabetic Eye Image Dataset (MEH-MIDAS) and Kaggle EyePACs.[4]

**Model fine-tuning/ training**

The fine-tuning/ training pipelines for ResNet50, SwinV2, ViT-base, and RETFound are shown in **Supplementary Figure 1**. We retained the architecture of all pre-trained models, modifying only the output layer to match the required number of classes—binary classification (i.e. presence of disease) for all tasks, except DR which was a five-class classification task (i.e. normal, mild, moderate, severe non-proliferative and proliferative DR). For image preprocessing, we detected the retinal mask using the Hough Circle Transform, and then cropped the mask region to reduce the influence of black background. For data augmentation, we applied random horizontal flipping, scaling, and rotation after resizing and cropping the fundus images (SwinV2: 256×256 pixels; all other models: 224×224 pixels). **Supplementary Text 1 in Appendix** provides further details on each model's architecture.

All the models were fine-tuned/ trained on their task-specific datasets using two distinct approaches **(Figure 1).** In the first approach, we varied the fine-tuning/ training sample sizes by using different proportions of the full dataset (e.g., 20%, 50%, and 100%). In the second approach, we used smaller fixed sample sizes in absolute number, while maintaining balanced class ratio. Specifically, for all binary detection tasks, we used 200, 100, and 50 images per disease class, resulting in final sample sizes of 400, 200 and 100 images, respectively. For the five-class DR severity detection task, we included 100 and 50 images per DR severity class, resulting in final sample sizes of 500 and 250 images, respectively.

During the training of the traditional models, all model layers were updated (i.e. trained end-to-end) with parameters optimization performed using the Adam optimizer. ResNet50, ViT-base, and SwinV2, were fine-tuned over 40 epochs with a batch size of 16, fixed learning rate of $1 \times 10^{-5}$, weight decay of $5 \times 10^{-4}$, and with a default drop path rate. There is no layer decay rate for the traditional models **(Supplementary Table 3).** To mitigate the effects of class imbalance on the model performance, we applied the square-root inverse (SQINV) re-weighting technique to all models during the fine-tuning process across all sample sizes. On the other hand, RETFound was fine-tuned over 50 epochs with a batch size of 16, a base learning rate of $5 \times 10^{-3}$, a weight decay of 0.05, a drop path rate of 0.2, and also a layer decay rate of 0.65. The model weights were saved at the epoch that yielded the highest AUC performance on the validation set, and were applied for internal and external testing subsequently. All models' fine-tuning were performed in PyTorch (Python v.3.8, torch v.1.7.0, torchvision v.0.8.0). An overview of the model architectures of the involved models is shown in **Supplementary Figure 2.** Further details on the resource utilization for the models' fine-tuning/training and inference are shown in **Supplementary Table 3 and 4**.

**Evaluation Metrics and Statistical Analysis**

For the detection of ocular and systemic disease tasks (excluding DR detection), we evaluated the models' performance using area under the receiver operating characteristic curve (AUC) and the maximum F1 score. For the five-class DR detection, we first calculated the class-specific AUC and maximum F1 score, followed by macro-average AUC and macro-average maximum F1 score. The 95% confidence intervals (CI) were computed using 2000 bootstrap iterations. Pairwise comparisons between RETFound and the three traditional model approaches were conducted using two-tailed Z-tests with Bonferroni correction.[30] Statistical significance was defined as a P-value (P) less than 0.05/3 (P=0.017), accounting for multiple pairwise comparisons between RETFound and the three traditional models.

**Role of the funding source**

The funder(s) of the study had no role in study design, data collection, data analysis, data interpretation, writing of the report, or the decision to submit for publication. All authors had full access to all the data in the study and had final responsibility for the decision to submit for publication.

**Results**

The demographic characteristics of all study datasets utilized are detailed in **Supplementary Table 2**. In brief, the SEED study consisted of participants aged 40.0-91.3 years of Malay (SIMES), Indian (SINDI), and Chinese (SCES) ethnicities. The CIEMS consisted of Indian participants aged 30-100 years. The BES and CUHK-STDR consisted of Chinese participants aged 50-93 years and 15.2-86.7 years, respectively. The SP2 study consisted of individuals aged 24.6-94.9 years of Chinese, Indian, Malay ethnicities. UKBB consisted of individuals aged 40 to 76 years from White Caucasian, African ancestry, Asian, and other ethnicities. Approximately half of the individuals in each dataset were females. For the open-source datasets, namely, APTOS-2019 (India), IDRiD (India), MESSIDOR-2 (France), ODIR-5K (China), PAPILA (Spain), and GAMMA (China), the countries of origin are known; however, detailed demographic information is unavailable.

**Detection of ocular diseases**

For detection of eye diseases, RETFound demonstrated largely comparable AUC performance to the traditional models, when fine-tuned on full datasets. This was consistently observed in internal and external test sets **(Figure 2, Supplementary Table 5–8).** When fine-tuned with reduced sample sizes of ≤400 images, comparable performances between RETFound and the traditional models persisted for detection of disease-related VI and visually significant cataract (**Figure 2, Supplementary Table 5 & 6)**. Maximum F1 scores for these two tasks supported these findings (**Supplementary Table 12 & 13)**.

In contrast, for DR detection, when fine-tuned with smaller datasets (≤500 images), RETFound consistently outperformed ResNet50 (AUCs ranging from 0.625–0.811) and ViT-Base (AUCs ranging from 0.617–0.911) (all P≤0.004), in both internal and external tests **(Supplementary Table 7)**. However, in these scenarios of reduced datasets, SwinV2's performance (AUCs ranging from 0.707–0.923) remained comparable to RETFound (all P≥0.67) in both internal and external tests. Maximum F1 scores for DR detection also mirrored these trends **(Supplementary Table 14).**

For glaucoma detection, when fine-tuned with smaller sample sizes (≤400 images), RETFound (AUCs ranging from 0.738–0.990) showed a distinct advantage over ResNet50 (AUCs ranging from 0.613-0.868, all P≤0.002 [except for one comparison, when fine-tuned on 100 images using GAMMA dataset]) in both internal and external test sets. In these scenarios, SwinV2 (AUCs ranging from 0.646–0.947) remained comparable to RETFound across the internal and external test sets **(Supplementary Table 8)**. Maximum F1 score for glaucoma detection largely aligned with these observations **(Supplementary Table 15).**

**Supplementary Tables 3 and 4** summarised the computational resource demands during fine-tuning and inference, using disease-related VI detection as an example. RETFound required significantly higher GPU memory during fine-tuning, with peak usage estimated at 12.3 GB, compared to traditional DL models (ResNet50: 2.3 GB, ViT-base: 3.4 GB, SwinV2: 11.4 GB). Additionally, RETFound exhibited slower inference speeds, processing 60.7 images per second, compared to traditional models such as ResNet50 (101.3 images per second), ViT-base (90.1 images per second), and SwinV2 (70.1 images per second). Similar trends in computational resource demands were observed across other tasks (**see Appendix**).

**Detection of systemic diseases**

When detecting systemic diseases, RETFound did not exhibit distinctly superior performance compared to traditional models when fine-tuned on full datasets **(Figure 3 and Supplementary Tables 9–12)**. However, its advantage became more evident with smaller sample sizes (≤400 images).

For example, in detecting diabetes across internal and external datasets, fine-tuning RETFound with 400 images resulted in AUCs ranging from 0.568–0.731, outperforming ResNet50, which achieved AUCs of 0.517–0.678 (all $P<0.001$). **(Supplementary Table 9)** Consistent trends were observed with 200 and 100 images, where RETFound maintained its superior performance over ResNet50. Additionally, when fine-tuned with 200 and 100 images, RETFound also demonstrated superior performance compared to ViT-base and SwinV2, achieving higher AUCs across most comparisons (all P≤0.016). Similarly, maximum F1 scores for diabetes detection indicated that RETFound consistently performed the best when fine-tuned on 200 images or fewer. **(Supplementary Table 16)**

For hypertension detection, when fine-tuned on full, 50%, or 20% datasets, RETFound did not show significant differences with the traditional models in internal tests. However, when fine-tuned with 400 images or lesser, across both internal and external test sets, RETFound showed the highest AUC. Notably, RETFound showed significantly better AUCs (0.730–0.740 for SP2; 0.599–0.622 for UKBB) than the other three traditional models (AUCs: 0.653–0.720 for SP2; 0.557–0.612 for UKBB; all P<0.001 except for two comparisons). **(Figure 3b & Supplementary Table 10)** Maximum F1 scores for hypertension detection mirrored these trends. **(Supplementary Table 17)**

For CKD detection, when fine-tuned on full or 50% datasets, RETFound showed no significant differences with the traditional models in internal tests. Nevertheless, across both internal and external test sets, fine-tuning RETFound with 20% of the data resulted in AUCs of 0.635–0.847, significantly outperforming ResNet50 (AUCs: 0.594-0.807, all P≤0.01). **(Supplementary Table 11)** Additionally, fine-tuning RETFound with 100 images resulted in AUCs of 0.635–0.832, outperforming ViT-base (AUCs: 0.601–0.759, all P≤0.008) and SwinV2 (AUCs: 0.527–0.778, all P≤0.005). **(Figure 3c, Supplementary Table 11)** Maximum F1 score for CKD detection consistently showed that RETFound performed the best when fine-tuned on 20% of datasets or lesser. **(Supplementary Table 18)**

**Discussion**

Our study evaluated and compared the performance, label efficiency, and computational resource demands of RETFound, a retina-specific FM, with commonly used traditional DL model approaches, including ResNet50, ViT-base, and SwinV2. These comparisons were conducted across various downstream ocular and systemic disease detection tasks, using varying fine-tuning sample sizes and multiple external test sets. For ocular disease detection, traditional DL models demonstrated largely comparable performance to RETFound, indicating traditional models continued usefulness in these tasks. On the contrary, for systemic disease detection, when fine-tuned on smaller datasets, RETFound largely outperformed traditional models. These findings provide valuable insights into the practical considerations of adopting RETFound over traditional models, accounting for performance, label efficiency, and computer resources. This comprehensive evaluation explicitly highlights scenarios where RETFound's pretraining approach offers tangible improvements in clinical performance and applications, as well as situations where traditional model remain sufficient and resource-efficient.

The comparable performances between RETFound and traditional models in ocular disease detection tasks, when fine-tuned on full datasets, suggest that in scenarios with abundant labeled data, conventional model approaches may suffice to achieve desired performance. This phenomenon may be attributed to the availability of sufficient data combined with the distinctiveness of disease features on retinal photos (particularly in major eye diseases). Notably, for detecting disease-related VI and cataract, this trend of comparable performance persisted even when fine-tuning on small sample sizes (200 and 100 images, with cases constituting half).

On the other hand, for DR and glaucoma detection, when the fine-tuning sample sizes were reduced to ≤100 images per class for DR or ≤400 images for glaucoma, RETFound outperformed ResNet50 in almost all test sets (**Figures 2c, 2d, and 4a**). However, SwinV2 remained largely comparable to RETFound even in scenarios with small sample sizes of 200 or 100 images (**Supplementary Tables 7 and 8**), indicating that the SwinV2 model may already be robust enough for detecting broad range of eye diseases. This suggests that SwinV2, a more advanced architecture than ResNet50[13], could offer a competitive alternative to RETFound in low-data settings for specific ocular disease detection tasks.

Additionally, in terms of computational resource consumption, RETFound required more GPU RAM during the fine-tuning phase and had a lower inference speed (i.e., processing fewer images in a given time) compared to the traditional DL models (**Supplementary Table 3, 4 and Appendix**). Moreover, the higher computational demands of RETFound may limit its feasibility and deployability in low-resource settings. Collectively, this indicates that traditional DL models remain a competitive and resource-efficient option for ocular disease detection tasks, especially for diseases characterized by distinct features presented on retinal photos or when large training datasets are available. These findings also highlight the importance of aligning model selection with available computing infrastructure in deployment sites, particularly in low-resource regions.

Conversely, RETFound's advantages became more apparent when detecting systemic diseases in scenarios with limited fine-tuning sample sizes. Across tasks involving systemic disease detection from retinal photos, RETFound consistently outperformed traditional DL models pre-trained on natural images, particularly with smaller fine-tuning sample sizes (**Supplementary Tables 9–11**). This enhanced performance can be attributed to RETFound's pretraining approach, which leveraged ~900,000 unlabeled retinal images. The SSL pretext task involved reconstructing highly masked retinal images, compelling the model to infer missing information from limited visible patches[4]. This process likely enabled RETFound to capture subtle, retina-specific contexts, such as fine-grained vascular patterns, which may be less discernible to the human eye or traditional models trained on natural images.[4,7] Overall, these findings underscore the potential additional value of pretraining on a retina-specific FM, for oculomics tasks (e.g., systemic disease detection). This phenomenon may be partly explained by the complexity and subtlety of retinal features that indicate systemic diseases, such as retinal vascular changes or features not yet known in current knowledge domain.[31-33]. These findings suggest that RETFound's superior ability to detect systemic diseases with limited data could help advance the development of oculomics algorithms, particularly in settings where well-annotated labeled datasets are scarce.

While RETFound demonstrated superior performance compared to traditional DL models in systemic disease detection, particularly with smaller fine-tuning sample sizes, it appeared to still fell short in fully addressing domain gaps, especially with cross-ethnic generalization during external validation. Among the three traditional DL models, we observed reduction in AUC performance in external test sets, especially in BES (China) and UKBB (United Kingdom), as compared to SP2 (Singapore) which has similar ethnic composition as the fine-tuning dataset, SEED (Singapore). However, even with RETFound, which was anticipated to exhibit better generalizability, significant drops in BES and UKBB were still observed (**Supplementary Tables 9–11**). A similar trend was observed in DR detection, where a larger drop in AUC was noted in MESSIDOR-2 (France) compared to IDRID (India) which likely have more similar ethnic composition as the fine-tuning dataset, APTOS-2019 (India). Similarly, in glaucoma detection, a larger drop in AUC was observed in PAPILA (Spain) compared to the two Asian external test sets, ODIR-5k (China) and GAMMA (China). These findings indicate the need for future research to address cross-ethnic generalization gaps by incorporating more diverse, multi-ethnic datasets during pretraining/training, thereby improving the robustness and equity of foundation models across populations.

**Strengths and limitations**

The strength of our study lies in its extensive and robust external testing, conducted using multiple test sets from diverse population cohort studies and open-source datasets. To better evaluate generalizability, in our study design, we incorporated three external test sets for each task. Second, we included a broad range of scenarios in our evaluations by using varying sample sizes for fine-tuning, both by proportions and absolute sizes. This approach is particularly valuable when assessing performance under different specific scenarios, including conditions of limited data availability. Third, while previous studies have utilized RETFound for ocular disease detection tasks[34,35], our study is among the first to comprehensively compare traditional DL model approaches with RETFound across both ocular and systemic disease detection tasks. By specifically evaluating the SSL-based RETFound against SL-ImageNet-based traditional DL models, our findings provided deeper understanding of the respective strengths and limitations of traditional models and retina-specific FMs.

Our study also has several limitations. First, while RETFound and the traditional models were evaluated for detecting a range of ocular and systemic diseases, longitudinal prediction tasks and the detection of rare ocular and systemic diseases were not included. To provide deeper insights into the comparative performance and broader applicability of these models, future research can further explore use cases such as the prediction of dementia, coronary artery disease, stroke, as well as detection of rare diseases. Second, as new backbone architectures and retina-, eye-specific FMs continue to advance[5,7,36,37], ongoing evaluation remains crucial. Finally, although we assessed computational resource demands, our current study's scope did not cover the environmental and economic costs of large pre-training FM. These critical aspects warrant future investigation to balance technological advancements with sustainability considerations.[38]

**Future consideration**

There are still a few areas that also warrant further exploration. First, it would be valuable to determine whether an optimal minimum fine-tuning sample size exists to achieve adequate performance, particularly when using FMs in data-scarce settings. Second, a deeper understanding of how demographic diversity and quality of training data influence downstream performance is crucial. Investigating the interrelationships between pre-training (e.g., identifying the extent to which pre-training data should be expanded to maximize generalizability), fine-tuning (e.g., assessing the impact of image quality and labeling accuracy), and test datasets could guide the development of more robust and generalizable models.[39,40] Third, future exploration on emerging multimodal FMs, which have the potential to learn diverse features from multiple imaging modalities[41], may yield better generalisability, new insights and uncover novel applications.

**Conclusion**

Traditional DL models are mostly comparable to RETFound for ocular disease detection, especially when trained on sufficiently large, labeled dataset. However, in scenarios of small datasets, RETFound demonstrated superior performance in systemic disease detection. These findings provide unique insights and balanced perspectives into the respective merits and limitations of traditional DL models and FM. Holistic considerations of practical factors, including performance, label efficiency, and computational resource requirements, should guide the selection of traditional models or FMs for specific clinical contexts.

Figure 1. Overview of study design.

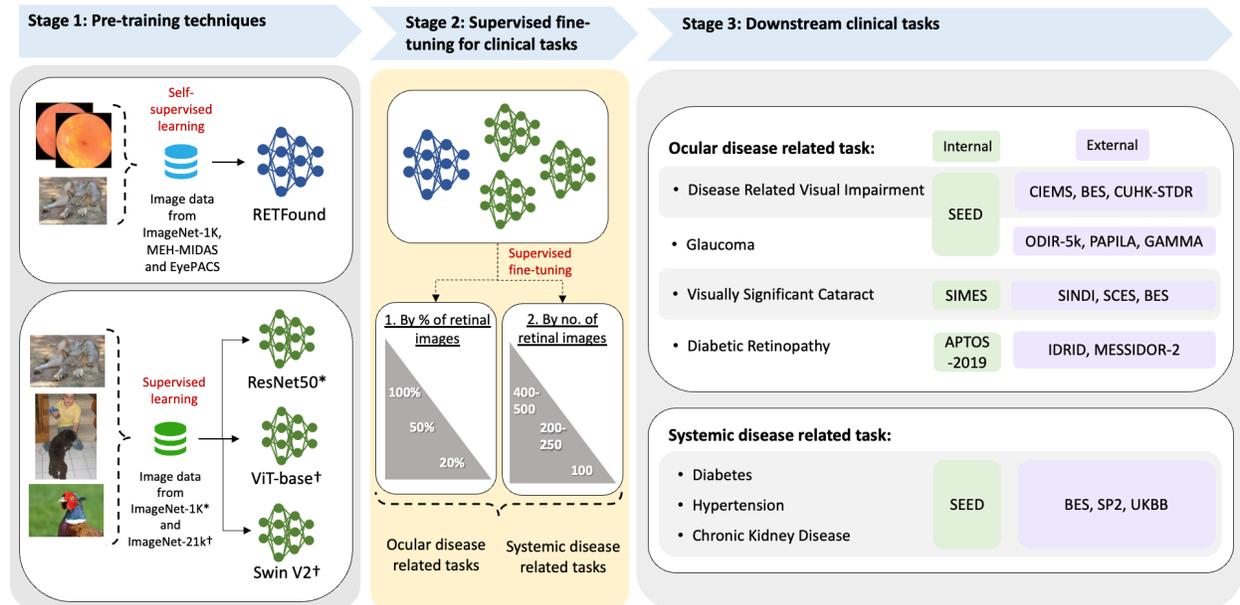

**\***=applies to ResNet50 model only. **†**=applies to ViT-base and SwinV2 models only. **MEH-MIDAS**= Moorfields Diabetic imAge dataset. **SEED**= Singapore Epidemiology of Eye Diseases. **CIEMS**= Central India Eye and Medical study. **BES**=Beijing Eye study. **CUHK-STDR**= Chinese University of Hong Kong's Sight Threatening Diabetic Retinopathy study. **ODIR-5k**= Ocular Disease Recognition-5k. **GAMMA**= Glaucoma grAding from Multi-Modality imAges. **SIMES=** Singapore Malay Eye Study. **SINDI**= Singapore Indian Eye Study. **SCES**= Singapore Chinese Eye Study. **APTOS-2019**= Asia Pacific Tele-Ophthalmology Society 2019. **IDRID**= Indian Diabetic Retinopathy Image Dataset. **MESSIDOR-2**= Methods to Evaluate Segmentation and Indexing Techniques in the field of Retinal Ophthalmology. **SP2=** Singapore Prospective Study. **UKBB**= United Kingdom Biobank.

Figure 2. Performance of the models in the internal test set, when fine-tuned on 100%, 50%, 200, and 100 images for ocular disease detection:
a) disease-related visual impairment, b) visually significant cataract, c) diabetic retinopathy (fine-tuned on 500 and 250 images), d) glaucoma.

**a)** Disease-related visual impairment

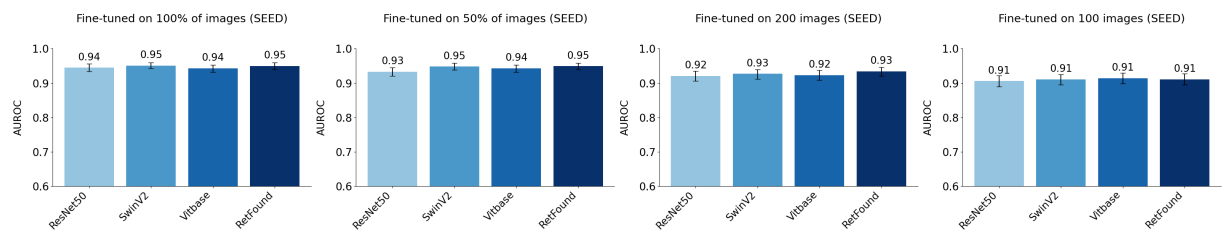

**b)** Visually Significant Cataract

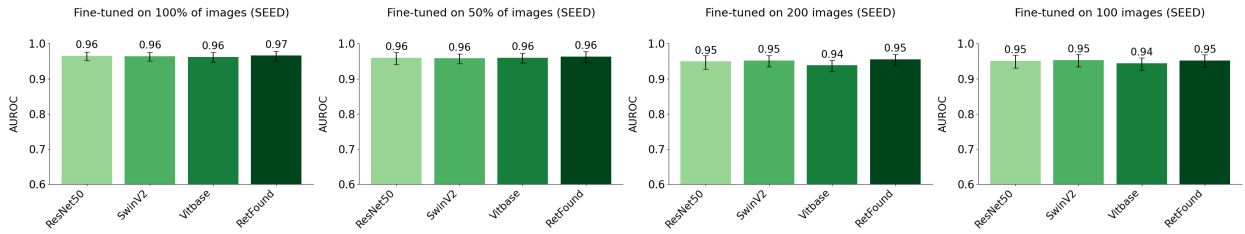

c) Diabetic Retinopathy

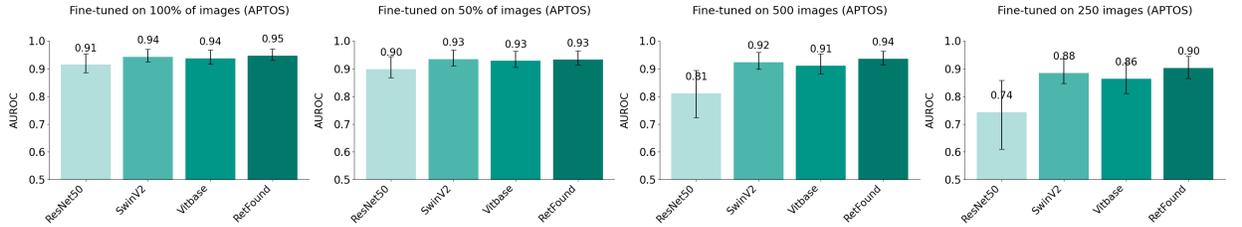

d) Glaucoma

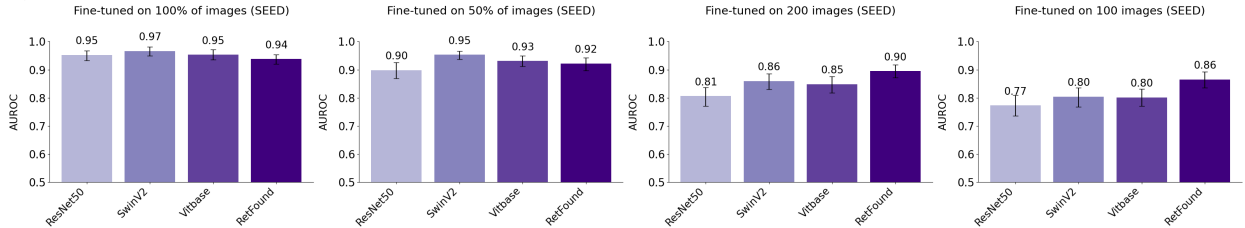

Figure 3. Performance of the models in the internal test set, when fine-tuned on 100%, 50%, 200 and 100 images for systemic disease detection:

a) diabetes, b) hypertension, and c) chronic kidney disease.

**a)** Diabetes

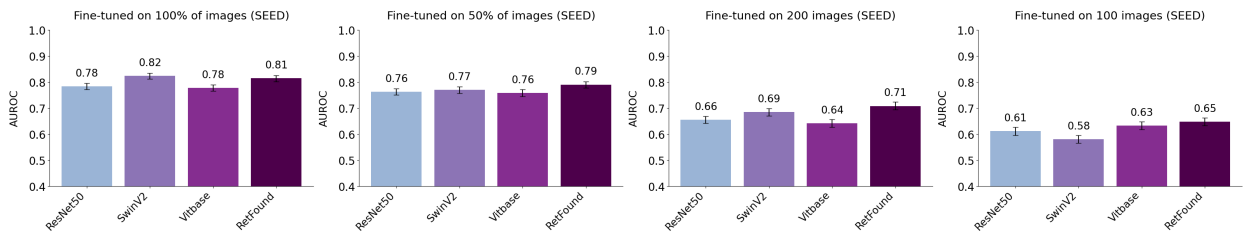

**b)** Hypertension

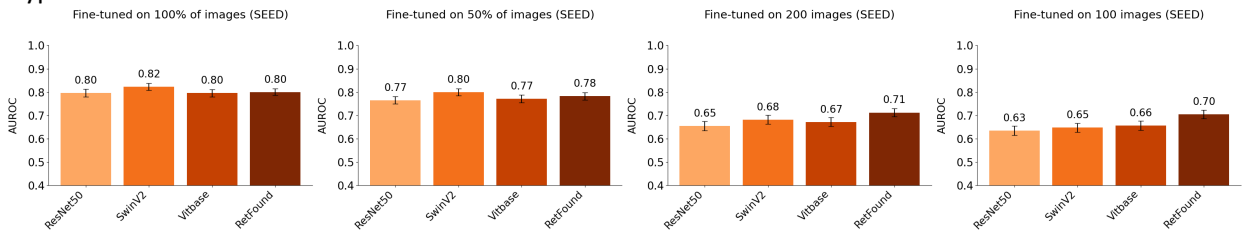

**c)** Chronic Kidney Disease

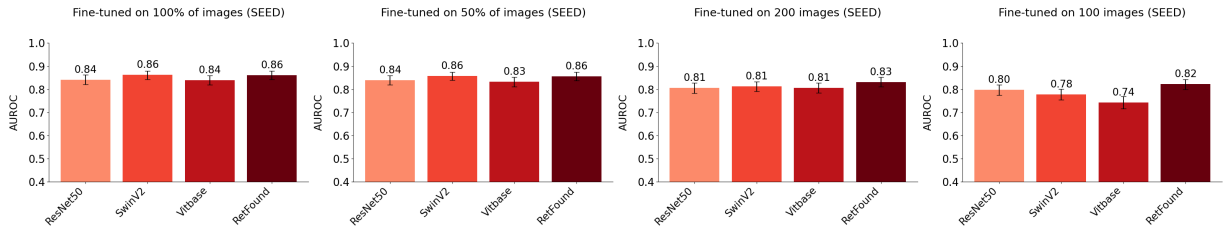

Figure 4a. AUC performance of RETFound, ResNet50, SwinV2, and ViT-base across varying fine-tuning dataset sizes for ocular disease detection, in internal tests.

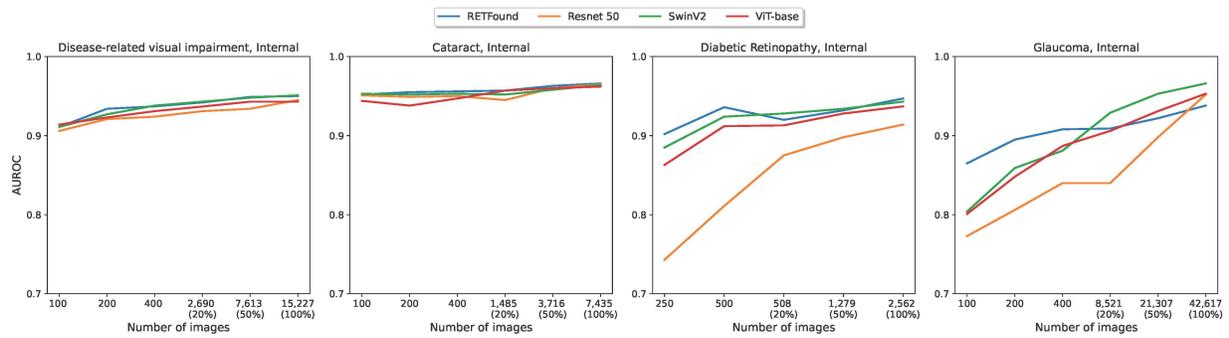

Figure 4b. AUC performance of RETFound, ResNet50, SwinV2, and ViT-base across varying fine-tuning dataset sizes for systemic disease detection, in internal tests.

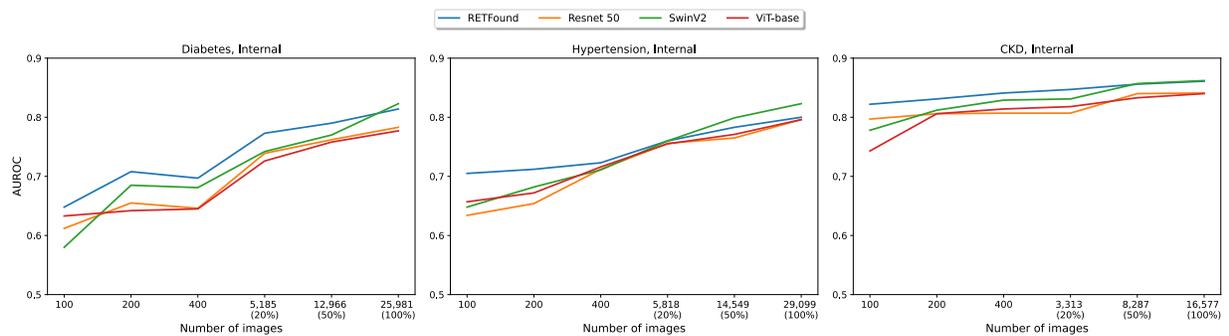